# Feature Vector Regularization in Machine Learning

Yue Fan[1], Mark Kon[1,2], and Louise Raphael[3]

**Abstract:** Problems in machine learning (ML) often involve noisy input data, and ML classification methods have in some cases reached limiting accuracies when they are based on standard ML data sets, e.g. consisting of feature vectors and classes. An important step toward greater accuracy in ML will require incorporation of prior structural information on data into learning. We will denote methods which regularize feature vectors as unsupervised regularization methods, analogous to supervised regularization methods used to estimate functions in machine learning. We study regularization (denoising) of ML feature vectors using analogues of Tikhonov and other regularization methods for functions on $\mathbb{R}^n$. A feature vector $\mathbf{x} = (x_1, \ldots, x_n) = \{x_q\}_{q=1}^n$ is viewed as a function of its index $q$, and smoothed using some prior information on the structure of the feature vector. This can involve a penalty functional on feature vectors analogous to those in statistical learning, or use of some proximity (e.g. graph) structure on the set of indices $q$ (the *index space*). Such feature vector regularization inherits a property from function denoising on $\mathbb{R}^n$, in that denoising accuracy is non-monotonic in the denoising (regularization) parameter $\alpha$. Under some assumptions about the noise level and the data structure, we show that the best reconstruction accuracy also occurs at a finite positive $\alpha$ in index spaces with graph structures. We adapt two standard function denoising methods used on $\mathbb{R}^n$, local averaging and kernel regression. In general the index space can be any discrete set with a notion of proximity, e.g. a metric space, a subset of $\mathbb{R}^n$, or a graph/network, with feature vectors as functions with some notion of continuity. We show this improves feature vector recovery, and thus the subsequent classification or regression done on them. We give an example in gene expression analysis for cancer classification, with the genome as an index space with a network structure based on prior knowledge of protein-protein interactions.

## 1. Introduction

Noise from irregular or imprecisely measured data is an important problem in machine learning, particularly in computational biology applications [1, 2]. A single gene expression array can measure human gene expressions in the tens of thousands. However, measurement errors and other sources of variation in gene expression can interfere with the learning and other algorithms that use them to classify cancer and other tissue subtypes [3-5]. The accuracies of regression and classification in such high dimensional problems are in some cases reaching limits, in particular when they are based strictly on information in datasets, without any known structural constraints. An important step in improving this could involve using prior knowledge of constraints in data structures, based on domain (e.g. biological) information [[6]; [7] [8]]. In comparison, the machine learning problem of regularizing and denoising functions defined on Euclidean spaces $\mathbb{R}^d$ (e.g. from visual images) has been studied very broadly ([9] [10] [11]). Regularization methods exploit continuity and other prior constraints on functions that are measured, and also the lack of such continuity or constraints on noise, which is largely stochastically independent. One denoising method is to average or smooth empirically measured functions over adjacent locations (e.g. pixels) in order to maximally quench noise (variance) and minimally add error to true function values (bias). Other regularization methods based on noisy as well as partial information about functions have been studied widely in ML, and these again involve regularization methods. For example, given a function $f(\mathbf{x})$ whose (noisy) values $y_i$ are known at points $\{\mathbf{x}_i\}_{i=1}^n$ in $\mathbb{R}^d$, a regularization method typically estimates $f$ as the minimizer of a *regularization functional*

$$L(f) = \sum |f(\mathbf{x}_i) - y_i|^2 + \alpha \|f\|^2 \tag{1.1}$$

where the second term $\alpha \|f\|^2$ represents an appropriate penalty representing the deviation of $f$ from known prior constraints, while the first term is a penalty for deviation of the approximating $f$ from the measured values at the points $\mathbf{x}_i \in \mathbb{R}^d$.

---

[1] Department of Mathematics and Statistics, Boston University, Boston, MA 02215
[2] Department of Mathematics and Center for Theoretical Physics, MIT, Cambridge, MA 02139
[3] Department of Mathematics, Howard University, Washington, DC, 20059

The success of such regularization methods suggests that similar methods might be useful in a different stage of the machine learning process, the formation of the feature vector $\mathbf{x} = (x_1, \ldots, x_n)$ itself. To exploit methods previously used for function regularization, we will view an ML feature vector $\mathbf{z} = (z_1, \ldots, z_n)$, e.g., defining expressions of genes $q = 1, \ldots, n$, as a *feature function* $z(q)$ defined on its *index* space (of genes) $q$, $1 \leq q \leq n$. This allows the adoption of well-studied machine learning and functional and numerical analysis tools for denoising and interpolation of functions on $\mathbb{R}^d$. Thus, the spatial structure of $\mathbb{R}^d$ can be replaced by a metric or network structure on the space $G$ of genes $q$. More generally the feature index $q$ can range over any structured set on which there is a prior notion of nearness, such as a metric or network structure. The resulting denoised feature vectors (e.g. gene arrays) can lead to more accurate predictions (for example of cancer subtypes). Regularization of a feature vector requires prior information, for example that (as function on $q$) it is 'continuous', meaning that 'nearby' indices $q$ yield 'nearby' features $z_q = z(q)$. In practice the space of $q$ can be a topological or metric space, the real line, or a graph/network. For example, averaging the value $z(q)$ at index $q$ with the values $z(r)$ of its neighbors will average out noise, reinforcing similarity and diminishing individually high errors. We will denote this type of regularization of feature vectors (without use of the information in function values $y_i$) as *unsupervised regularization.*

An example of such regularization in numerical analysis occurs in processing of very noisy images, where blurring (softening) an image improves visual information content. This assumes adjacent pixels (feature indices) have nearby illumination levels (feature values). The bias from blurring is more than offset by the variance reduction in noise averaging. There is an optimal amount of blurring – too little or too much will diminish the information content. This can apply to many types feature vectors structured with continuity relationships between adjacent features, i.e., with features $z_q$ similar if their indices $q$ are nearby.

As seen in blurring (and denoising on $\mathbb{R}^n$), reconstruction accuracy can be non-monotonic, first increasing and then decreasing in the regularization parameter, so the best accuracy occurs at some (but not too much) regularization. Here we will state and prove sufficient conditions for this to occur for unsupervised regularization ML index spaces $G$ with network structures. Thus we seek conditions under which regularization can be shown to help. In supervised machine learning this non-monotonicity in the regularization parameter $\alpha$ is handled through optimization of $\alpha$ using cross-validation and other structured risk minimization methods ([6, 7]). In the unsupervised context of feature vector regularization, this must be handled differently, since accuracy of the regularization cannot be measured. One way to optimize $\alpha$ is to measure the performance of the unsupervised regularization method in a subsequent supervised learning task (see Section 6).

Regularization approaches for functions on $\mathbb{R}^d$ have been studied widely in the functional and numerical analysis literature. Explicit attention to optimal regularization was introduced originally by Tikhonov [12]. In Tikhonov regularization, there is partial information about a function $f(x)$, and the ill posed problem of inverting this to a unique estimate of $f$ is solved by adding a regularization requirement that some functional or norm $\|f\|$ be minimized (see (1.1) above). This is the basis regularization methods in statistical machine learning ([6, 7]).

These ideas were extended by Krukovskii [13], who showed that if $f(x)$ is perturbed by noise $\eta(x)$, then a regularization $f_{1\alpha}$ of $f_1(x) = f(x) + \eta(x)$ (with regularization parameter $\alpha$ scaled properly in the magnitude $\|\eta\|$ of the noise) can recover $f(x)$ from pointwise errors better than direct measurement of $f_1$. This was made precise in an asymptotic result as both $\|\eta\|$ and $\alpha$ go to 0 in a scaled way, and illustrated the power of regularization against noise in numerical analysis. Our results, on the other hand, show when such regularization (on discrete spaces) can help non-asymptotically. We give conditions when, for fixed noise $\eta$ and regularization $\alpha$, the approximations $f_{1\alpha}$ improve over direct measurement of $f_1$, i.e., when regularization cancels noise in $f_1 = f + \eta$. This is not novel in numerical analysis, but it has not been applied in general to regularizing feature vectors.

**Regularization of feature vectors.** Regularization denoising of feature vectors is thus pre-processing step *prior to* any supervised learning of the data, for example using a classification algorithm. We will illustrate this in section 6 for regularization of gene expression vectors used for predicting cancer metastasis. In particular our approach is not aimed at providing better machine learning decision/discrimination rules,- these are typically downstream from unsupervised regularization. This viewpoint has effectively been used previously e.g., in computational biology applications, where prior group, network or metric structures on gene sets have been used to regularize gene expression information [3, 14].

**Prior Work.** Denoising of gene expression arrays based on gene networks has been used elsewhere, including [3], in which local gene clusters based on a protein-protein interaction (PPI) network were used to provide averaged features for discrimination in cancer prognosis. Unlike the approach here, where preprocessing is unsupervised (i.e., does not use the classes of the samples), these clusters are built up in a supervised way so as to maximally differentiate cases and controls, in the same cancer metastasis data sets (of Wang [5] and van de Vijver [4]) analyzed by us.

Local averaging in gene networks has been used in other contexts. Kasif, et al. [15] uses biochemical pathway and more generally gene network-based averaging to strengthen signals in gene expression arrays. That method, GNEA (Gene Network Enrichment Analysis) stabilizes gene expression signals by averaging network-connected expressions. Hammond, et al. [16] studies localized wavelets on graphs based on spectral decomposition of the graph Laplacian, yielding canonical (graph-based) clustering methods alternative to those used here.

Methods for regularizing (smoothing) functions on graphs using spectral approaches have also been studied recently. Smola and Kondor [17] studied Laplacian eigenfunctions on graphs to regularize graph-based functions. Decomposing a function $f$ on the graph in eigenfunctions of the graph Laplacian gives compression and dimensional reduction via projection onto a few Laplacian eigenfunctions. Since the eigenfunctions typically have local support, this can also decompose $f$ into canonical local clusters.

In Rapaport, et al., [18], spectral denoising techniques for functions on graphs are used to regularize gene expression arrays also viewed as functions on the gene network. Higher spectral components of expression functions are eliminated in a smoothing process giving less noisy arrays for prediction and classification.

Szlam, et al. [19] overviews diffusions on graphs as means of smoothing (generally noisy) functions on them. Belkin [20] also studies dimensional reduction via regularization on graphs using the graph Laplacian and the heat equation. Bougleux, et al. [21] study alternative methods for function regularization on graphs using energy functional minimization. This is used on image functions in which pixels come with an adjacency structure, using a procedure analogous to smoothing methods for functions on $\mathbb{R}^d$.

**Local averaging and kernel regression.** Classical methods for denoising Euclidean functions include local averaging [22, 23]; convolution methods [24]; and support vector regression [25]. Our examples will first generalize local averaging; a prototypical example for this on $\mathbb{R}^d$ is Haar wavelet denoising [22, 23]). Finite local averages form a finite martingale of approximations of the original feature function $f_1(q)$ on the (index) space $G$ of indices $q$. The sequence, indexed by a parameter $t$, consists of locally averaged conditional expectations $f_{1t}$ of the perturbed ideal feature function $f$, given as $f_1 = f + \eta$, with $\eta$ representing noise and batch effects. Here increasing $t$ represents averages over larger local sets of indices $q$.

The sequence $f_{1t}$ of approximations, like its analog in numerical denoising, approximates the 'true' feature function $f(q)$ with an accuracy that increases and then decreases, in $t$. This maximum in accuracy (Theorems 1 and 3) represents a balance in bias versus variance [26-29], so competing errors due to bias (over-averaging) and variance (under-quenching of noise) balance at the optimal regularization level (with minimal error $\| f_{1t} - f \|$).

In Theorem 1 we give conditions for this regularization to help for function denoising on $\mathbb{R}$, and then for denoising feature vectors on graph structures (Theorem 3). We show when such regularization of feature vectors $\mathbf{x} = (x_i)_{i=1}^n = x(q)$ (through averaging using the graph structure of $q \in G$) improves over no regularization. In section 6 we numerically study this non-monotonicity of error for several gene expression datasets, demonstrating optimal accuracy for intermediate clustering levels $t$.

We average feature values $z(q)$ over different hierarchical clusterings of graph indices, with indices $q$ and $r$ clustered together based on graph proximity. These clusters are hierarchical, with the clustering (partition) $A_t$ at time $t_1$ is a sub-clustering of that at time $t_2 > t_1$. Equivalently, $\sigma$-field $\mathcal{F}_t$ generated by the partition $A_t$ forms a filtration, i.e., $\mathcal{F}_{t_1} \subseteq \mathcal{F}_{t_2}$ for $t_1 \leq t_2$. Thus on a large graph $G$, we consider a filtration $\mathcal{F}_t$ giving local cluster averages of the feature function $f(q)$ as a conditional expectation, $f_{1t} = E(f_1 \| \mathcal{F}_t)$, which forms a martingale ([30]). With some uniformity assumptions on the filtering $\mathcal{F}_t$ (see Section 3, Theorem 3) we have:

**Theorem 3**: Let $F$ be a space of feature vectors (feature space) with a basis $\{b_q\}_{q \in G}$ indexed by a graph $G$, and let $f_1 = f + \eta \in F$ be a noisy feature vector, with independent Gaussian noise $\eta \sim N(0, \varepsilon I)$. Let $\{\mathcal{F}_t\}_{0 \leq t \leq T}$ be a filter of $G$, with $\sigma$-fields $\mathcal{F}_{t_1} \subseteq \mathcal{F}_{t_2}$ for $t_1 \leq t_2$. Then the error of the averaging regularization $f_{1t} = E(f + \eta \| \mathcal{F}_t)$ over $0 < t < T$ is non-monotonic, i.e., there is a $0 < t < T$ for which the regularization is optimal (minimizes $\| f - f_{1t} \|$ in $L^2$ norm). In particular the optimal regularization is non-trivial, so $t > 0$. This statement is uniform over all graphs $G$, chains $\{A_t\}_{0 \leq t \leq T}$ of refinements and functions $f$ on $G$.

Our second approach, also adapted from numerical analysis adapts the function denoising technique of kernel regression into a regularization method for feature vectors. This is done again using graph structures on their index sets.

**Theorem 5**: Let $G$ be a finite graph, and $f(q)$ a function on $G$. Define the noisy function $f_1(q) = f(q) + \epsilon g(q)$, with $g(q)$ standard Gaussian noise on $G$. Let $K_\alpha(q, q_1)$ be a kernel converging to the identity as $\alpha \to 0$. Then for every sufficiently small noise level $\varepsilon > 0$, the expected error $E(\| f - f_{1\alpha} \|)$ of the kernel regression approximation $f_{1\alpha}(q) = \sum_G f_1(q_1) K_\alpha(q_1, q) dq_1$ is minimized at a positive value of the regularization parameter $\alpha$, so this regularization improves feature vectors based on $G$.

In numerical analysis, an alternative method to kernel regression is direct convolution of the signal function $f(q)$ with a denoising kernel $g(q)$ (such as a Gaussian), creating a smoothed signal more stable with respect to fluctuations ([31]). Direct convolution (not studied here) uses a pre-defined convolution kernel, while kernel regression uses a kernel solving an optimization problem.

**Application.** We will illustrate such denoising on gene expression arrays, as expression functions on gene networks,. This will be done using cluster-based local averaging and will numerically illustrate the above martingales and the non-monotonicity of the error in clustering parameter *t*, showing error that has a global minimum at intermediate values.

We will also look at network-based support vector regression, which convolves a Gaussian $g(q)$ with the expression function $f(q)$, creating smoothed feature vectors more stable against fluctuations. This is illustrated (as a preprocessing step) on gene expression arrays for predicting breast cancer metastasis. The network structure gives a prior notion of distance on the feature index set, which admits kernel regression on expression vectors.

We use two benchmark gene expression cancer datasets, of Van de Vijver [4] and Wang [5], both analyzing gene expression against metastasis outcomes in breast cancer. An announcement of these results was presented in [32]. In this paper we give complete proofs and provide tables of the numerical examples.

To order the topics in the paper, in Section 2 we begin by presenting the real line analog of our main graph/network theorem on local averaging. It shows that for optimized regularization of noisy functions on $\mathbb{R}^1$, the bias-variance tradeoff requires that the amount of averaging needed for optimal recovery is determined by noise to signal ratio, paralleling results like that of Krukowskii [13] mentioned earlier. Section 3 presents the analogous theorem for functions on graphs or networks. Sections 4 and 5 similarly prove theorems on kernel regression on Euclidean spaces, then extended to graphs.

Section 6 shows cancer gene expression arrays (as feature vectors) can be denoised this way, using protein interaction networks. With expression vectors smoothed on an appropriate network distance scale (similarity measure) they improve prediction of biological properties, here the prospective metastasis of given cancer.

The reading of this paper can be done either serially through the sections, or more compactly by reading this introduction and then moving directly to section 6.

## 2. Local averaging for denoising on $\mathbb{R}$

Local averaging methods for denoising functions $f(q)$ on $\mathbb{R}^d$ rely on the metric structure of $\mathbb{R}^d$, and assumptions on the continuity or smoothness of $f$. One family of such denoising methods (e.g. Haar wavelet denoising) uses averaging over local subsets of $\mathbb{R}^d$ which we call *clusters*. This method on $\mathbb{R}^d$ can be adapted to regularize or smooth functions on a graph $G$. In both cases the cluster-based averages can be viewed as conditional expectations.

On $\mathbb{R}^d$ consider a function $f_1(q) = f(q) + \eta(q)$, where $f$ is a signal and $\eta$ is noise. Let $t$ be a parameter, and $\{A_t\}_{t \in \tau}$ be a sequence of clusterings (i.e. for each $t$, $A_t$ is a collection of disjoint sets partitioning $\mathbb{R}^d$). We assume the sequence is nested, so that for $t_1 < t_2$, clustering $A_{t_2}$ is the same as or a refinement of $A_{t_1}$.

We view these clusterings on $\mathbb{R}^d$ in terms of their $\sigma$-fields $\mathcal{F}_t = \sigma(A_t)$, with $\sigma(A_t)$ the $\sigma$-field (minimal collection of subsets closed under complements and countable unions) generated by partition $A_t$. The collection $\{\mathcal{F}_t\}_{t>0}$ is a nested sequence of $\sigma$-fields, or a *filtration* on $\mathbb{R}^d$. The function $f_{1t}$ formed by averaging $f_1$ over clusters $A_t$ is the conditional expectation $f_{1t} = E(f_1 \| \mathcal{F}_t)$.

Mathematically the conditional expectations
$$f_{1t} = E(f_1 \| \mathcal{F}_t) = E(f \| \mathcal{F}_t) + E(\eta \| \mathcal{F}_t)$$
that average $f_1 = f + \eta$ over the clusters of $\mathcal{F}_t$ form a martingale, a collection $\{f_{1t}\}_{t \in \tau}$ of increasingly informative functions (with increasing $t$), with $f_{1t}$ measurable with respect to $\mathcal{F}_t$. Though there is successively more information in $f_{1t}(q)$ with increasing $t$, there is a local maximum in the accuracy of $f_{1t}$ in reconstructing the signal $f(q)$, at a point $t = t_0$. This occurs where the averaged noise $\eta_{t_0} = E(\eta \| \mathcal{F}_{t_0})$ in $f_{1t_0}$ is optimally quenched by cluster averaging in $\mathcal{F}_{t_0}$ (variance is reduced), without sacrificing the information content (addition of too much bias from over-averaging $f$ in $f_t$). For $t < t_0$ decreasing away from $t_0$ the function $f_{1t} = f_t + \eta_t$ becomes less informative, since noise $\eta_t$ exceeds the signal $f_t$ without significant averaging over nearby values. For $t > t_0$ increasing, information is lost, on the other hand, because over-averaging in $f_t$ on large clusters diminishes the original signal in $f$, biasing it.

We study this here for functions on $\mathbb{R}^n$ and then on networks, and will also illustrate an application to gene expression arrays. We start with an example theorem on $\mathbb{R}$ for denoising continuous $f(q)$. We will repeat for completeness some definitions in martingale theory.

**Definition 1**: For a domain $G \subset \mathbb{R}^n$, let $\mathcal{F}$ be the collection of Borel sets, i.e. the minimal extension of the collection of open sets that is closed under complements and countable unions and forms a $\sigma$-field. A *filter* of $\mathcal{F}$ on $G$ is an increasing sequence $\mathcal{F}_t$ of $\sigma$-fields (so $\mathcal{F}_{t_1} \subset \mathcal{F}_{t_2}$ if $t_2 > t_1$) on $G$ indexed by a non-negative integer parameter $t$, $0 \le t \le T$, with $\bigcup_t \mathcal{F}_t = \mathcal{F}$. For fixed $t$, the *clustering* $A_t$ of $G$ based on $\{\mathcal{F}_t\}_t$ is the most refined partition of $G$ measurable with respect to $\mathcal{F}_t$, i.e., it is the maximal disjoint collection of sets in $\mathcal{F}_t$ such that all sets in $\mathcal{F}_t$ are unions of sets in $A_t$.

Let $\mathcal{F}_t$ be a filter of $\mathcal{F}$ on $G$. Recall that in case the $\sigma$-field $\mathcal{F}_t$ is discrete (i.e., is composed of a finite number of sets) then $f_{1t} = E(f \| \mathcal{F}_{1t})$ consists of $f_1$ averaged over sets in the clustering $A_t$ generated by $\mathcal{F}_t \subset \mathcal{F}$. For example, if $\mathcal{F}_t$ is finite (in addition to being discrete) then
$$f_{1t}(q) = \frac{1}{|a|} \sum_{q \in a} f_1(q) \quad (\text{for } q \in a \in A_t),$$

where $|a|$ is the cardinality of set $a \in A_t$. This definition on the real line (again in case $\mathcal{F}$ is finite) is similar: for $q \in a \in A_t$ we define $f_{1t}(q)$ to be the average of $f_1(q)$ (now an integral) over set $a$.

**Definition 2:** The above successive averaging of a function over successive $\sigma$-fields is called the *martingale on G* defined by the function *f* and the filter $\mathcal{F}_t$.

The martingale $f_t = E(f \| \mathcal{F}_t)$ (of approximations to the underlying function $f$) is increasingly informative about *f* as *t* increases. Indeed, $f_t$ is an orthogonal projection of $f$ onto the increasing family of spaces $L^2(\mathcal{F}_t)$ (the square integrable functions measureable with respect to $\mathcal{F}_t$). Thus $\| f - f_t \|_2$ decreases monotonically for all $t$. This stops holding once noise is added to *f*. If we measure the noisy function $f_1(q) = f(q) + \eta(q)$ (with added mean zero random noise $\eta(q)$), then as *t* increases the conditional expectations $f_{1t} = E(f_1 \| \mathcal{F}_t)$ first give better approximations of *f*, which then can become successively worse for larger *t*. This is a case of too much information as *t* increases, in that incrementally more noise than signal is captured in $f_{1t}$.

It is well known that averaging out noise can improve the estimate of *f*, something that occurs in denoising using Haar wavelets (e.g., [23]). This can be illustrated on the unit interval with a continuous signal $f(q)$, $q \in [0,1]$, and additive noise $\eta(q) = \epsilon g(q)$, with $g(q)$ the standard Gaussian noise distribution (i.e., $\int_0^s g(q) dq = b(s) = N(0, \sqrt{s})$ is Brownian motion), with $\epsilon > 0$. Let the parameter $t = n$ vary through the positive integers, and $\mathcal{F}_t$ be the $\sigma$-field consisting of unions of sets $\left[\frac{i-1}{2^t}, \frac{i}{2^t}\right]$ for $(1 \leq i \leq 2^t)$. Consider the behavior (as $t \to \infty$) of the $L^2$ error between the signal $f(q)$ and its denoised approximation $f_{1t}(q) = E(f_1 \| \mathcal{F}_t)$ given by

$$R(t)^2 = \| f - f_{1t} \|_2^2 = \int_0^1 (f(q) - f_{1t}(q))^2 dq = \int_0^1 (f_t(q) + \eta_t(q) - f(q))^2 dq \,. (2.1)$$

We will need the fact that the noise-free error function $R_0(t) = \| f - f_t \|_2$ is strictly decreasing in *t* for any continuous non-constant *f* on $[0,1]$, using standard facts from the Haar wavelet approximation. Indeed, let $[a,b]$ be any interval on which *f* is continuous and non-constant, and $\overline{f}_0(x)$ be the constant function averaging $f$ on $[a,b]$. It follows easily that if $\overline{f}_1(x)$ is constant the left and right halves of $[a,b]$ (averaging $f$ on each half) then $\| f - \overline{f}_0 \|_2 > \| f - \overline{f}_1 \|_2$. [L,] The following theorem, stated on $[0,1]$, holds on any interval of the real line, and generalizes this fact.

**Theorem 1:** *If f is non-constant and continuously differentiable on the unit interval, then with probability arbitrarily close to 1 for sufficiently small noise level $\epsilon > 0$, the error $R(t) = \| f - f_{1t} \|_2$ is monotonically decreasing for sufficiently small t, and monotonically increasing for sufficiently large t, so it is minimized at an intermediate value $0 < t_0 < \infty$.*

That is, for $\delta > 0$ there exists a noise threshold $\epsilon > 0$ below which the error minimization statement above holds with probability greater than $1-\delta$. By monotonically decreasing we mean $E(R(t_1)) \leq E(R(t_2))$ for $t_1 \geq t_2$, with a similar statement for monotonically increasing.

**Sketch of Proof:** Note that for $t$ a nonnegative integer and $q \in \left(\frac{i-1}{2^t}, \frac{i}{2^t}\right] \subset [0,1]$, the function $g_t(q) = E(g(q) \| \mathcal{F}_t)$ takes the constant value

$$g_t(q) = 2^t \int_{(i-1)/2^t}^{i/2^t} dq' \, g(q') = 2^t 2^{-t/2} Z_i = 2^{t/2} Z_i,$$

with $Z_i$ independent and identically distributed (iid) standard normal random variables. Note that $R(t) = \| f_t(q) + \epsilon g_t(q) - f(q) \|_2$, so for sufficiently small $\epsilon > 0$, the contribution of noise $\epsilon g_t(q)$ to $R(t)$ is clearly negligible compared to that of the monotonically decreasing function $R_0(t) = \| f_t(q) - f(q) \|_2$. Thus it will follow (see below) that for fixed $t_1$ if $\epsilon$ is sufficiently small, $R(t)$ is monotonically decreasing in $t$ for all $t \leq t_1$ with probability greater than $1-\delta$, with $\lim_{\epsilon \to 0} \delta = 0$.

Note first that (all norms are in $L^2$)

$$R(t)^2 = \| f_{1t} - f \|^2 = \| f_t + \epsilon g_t(q) - f \|^2 = \int_0^1 dq \left[ (f_t - f)^2 + 2\epsilon g_t(f_t - f) + \epsilon^2 g_t^2 \right]$$

$$= \int_0^1 dq \left[ (f_t - f)^2 + \epsilon^2 g_t^2 \right] = \| f_t - f \|^2 + \epsilon^2 \| g_t \|^2,$$

using the fact that on each sub-interval $\left[\frac{i-1}{2^t}, \frac{i}{2^t}\right]$, the function $g_t$ is constant, while $\int_{(i-1)/2^t}^{i/2^t} dq (f_t - f) = 0$. Note that

$$\| g_t(q) \|^2 = \int_0^1 dq \, g_t^2(q) = \sum_{i=1}^{2^t} \int_{(i-1)/2^t}^{i/2^t} dq \, g_t^2 = \sum_{i=1}^{2^t} 2^{-t} \left( 2^{t/2} Z_i \right)^2 = \chi^2_{2^t}.$$

where $Z_i$ are iid $N(0,1)$ random variables and $\chi^2_n$ denotes a chi-square random variable with $n$ degrees of freedom. Hence

$$R(t)^2 = \| f_t - f \|^2 + \epsilon^2 \chi^2_{2^t} \tag{2.2}$$

Clearly, for $\epsilon$ sufficiently small, $R(t)$ is decreasing for small $t$ with probability arbitrarily close to 1, since $\| f_t - f \|$ is decreasing.

We now consider the case of large $t$. We first note that successive differences in $\| f_t - f \|$ converge to 0 as the integer $t \to \infty$. Thus successive differences in the sequence $R(t)$ are eventually bounded below by $R(t+1) - R(t) \geq -1 + \epsilon^2 (\chi^2_{2^{t+1}} - \chi^2_{2^t})$, with the two $\chi^2$ distributions defined by (2.2) above and not generally independent. In addition, the probability that the above term $-1 + \epsilon^2 (\chi^2_{2^{t+1}} - \chi^2_{2^t})$ is negative infinitely often is 0. This follows from the Borel-Cantelli lemma and (the sums below are all in $t$ ranging over integers)

$$\sum_{2^t>1/\epsilon^2}^{\infty} P\left(\chi^2_{2^{t+1}} - \chi^2_{2^t} - \frac{1}{\epsilon^2} < 0\right) \leq \sum_{2^t>1/\epsilon^2}^{\infty} P\left(\left|\left(\chi^2_{2^{t+1}} - \chi^2_{2^t} - \frac{1}{\epsilon^2}\right) - \left(2^{t+1} - 2^t - \frac{1}{\epsilon^2}\right)\right| > \left(2^{t+1} - 2^t - \frac{1}{\epsilon^2}\right)\right)$$

$$= \sum_{2^t>1/\epsilon^2}^{\infty} P\left(|(\chi^2_{2^{t+1}} - \chi^2_{2^t}) - 2^t| > 2^t - \frac{1}{\epsilon^2}\right) \leq \sum_{2^t>1/\epsilon^2}^{\infty} \frac{V(\chi^2_{2^{t+1}} - \chi^2_{2^t})}{\left(2^t - \frac{1}{\epsilon^2}\right)^2}$$

$$\leq \sum_{2^t>1/\epsilon^2}^{\infty} \frac{V(\chi^2_{2^{t+1}}) + V(\chi^2_{2^t}) + 2\sqrt{V(\chi^2_{2^{t+1}})V(\chi^2_{2^t})}}{\left(2^t - \frac{1}{\epsilon^2}\right)^2} = \sum_{2^t>1/\epsilon^2}^{\infty} \frac{2 \cdot 2^{t+1} + 2 \cdot 2^t + 2 \cdot \sqrt{2 \cdot 2^{t+1} \cdot 2 \cdot 2^t}}{\left(2^t - \frac{1}{\epsilon^2}\right)^2} < \infty,$$

where we have used Chebyshev's inequality. Note that the last inequality above is independent of the covariance of the two (dependent) $\chi^2$ distributions, and follows from the Schwartz inequality $|\text{cov}(A, B)| \leq \sqrt{V(A)V(B)}$. This shows that with probability 1, $R(t)$ is increasing with $t$ for $t$ sufficiently large, and completes the proof.

## 3. Graph martingale approximations

We will extend the above theorem for local averaging on $\mathbb{R}^1$, to clustering-based averages of functions on graphs, again defined as martingales. Since a graph/network structure $G$ is not necessarily consistent with a metric on $G$, the previous theorem on $\mathbb{R}$ does not directly extend to graphs or networks. Nevertheless the notion that proximity-based averaging can regularize noise carries over to this case. If network-based data are given as a noisy function $f_1(q)$ on a network $G$, then the network structure and an analog of continuity for $f_1$ can help eliminate noise, based on a graph theorem analogous to the above theorem on $\mathbb{R}^1$.

As an example, assume $G$ is a network of genes, and that the function $f_1(q) = f(q) + \eta(q)$ represents experimentally measured expression of gene $q \in G$ (here $q \in G$, $f(q)$ is the expression signal and $\eta$ noise). The network structure on $G$ is assumed to reflect biological relationships among genes that tend to make their expressions similar, so network-based clusters contain genes with a priori similar gene expressions. Thus cluster-averaged expressions, while biasing individual gene expressions, will quench noise through averaging over genes. Thus the bias is tolerable assuming gene expression $f(q)$ is 'continuous' on $G$.

**Definition 3.** A *graph* (or *network*) $\{G, w\}$ is a collection $G$ of *vertices*, with the *edges* defined as unordered pairs $(i, j)$. A symmetric function $w(i, j)$ $(i, j \in G)$, defined on the edges, has values known as the *weights* of $G$.

Let $\{\mathcal{F}_t\}_{t=0}^T$ be a filter on a finite graph $G$, with $\mathcal{F}_0 = \{G, \phi\}$ (the trivial $\sigma$-field) and $\mathcal{F}_T = 2^G$, $\sigma$-field of all subsets, with $t = 0, ..., T$ integer-valued.

Given a function $f(q)$ on $G$, consider the finite martingale on $G$ consisting of local averages $f_t = E(f \| \mathcal{F}_t)$ of $f$ over the clusters $A_t$ defined by $\mathcal{F}_t$. The measurements of $f$ are subject to noise $\eta(q) = \epsilon g(q)$. We assume (discretized) white noise, with $g(q)$ an $N(0,1)$ standard normal rv (random variable) for $q \in G$. We wish to approximate $f$ from its noisy measurements $f_1(q) = f(q) + \epsilon g(q)$, by projecting $f_1(q)$ onto its conditional expectation $f_{1t} = E(f_1 \| \mathcal{F}_t)$. We will show that for certain ranges of the parameters, the non-monotonicity of the error in $\varepsilon$ on the real line also occurs here. In what follows all function norms are $L^2$ norms on $G$ unless otherwise specified, and $V$, $E$ are variance and expectation respectively.

**Lemma 2:** On graph $G$, let $\{\mathcal{F}_t\}_{1 \le t \le T}$ be a filter with clustering $A_t$, and with $k_t \equiv |A_t|$ the number of clusters at level $t$. Assume $|A_{t+1}| \ge C |A_t|$ for a fixed $C > 1$. Letting $\eta_t(q) = E(\eta(q) \| \mathcal{F}_t)$, then uniformly over all graphs $G$ (and of course over $k_t$)

$$\frac{1}{\epsilon} E(\| \eta_t \|) = \sqrt{k_t} + O(1/\sqrt{k_t}) \qquad (k_t \to \infty).$$

and

$$\frac{1}{\epsilon^2} V(\| \eta_t \|) = O(1), \tag{3.1}$$

where $V$ is the variance.

*Proof:* We first estimate (below $a_t \in A_t$ is a cluster in clustering $A_t$)

$$\frac{1}{\epsilon} E(\| \eta_t \|_2) = E\left(\sqrt{\sum_{a \in A_t} \sum_{q \in a} \overline{g}_a(q)^2}\right) = E\left(\sqrt{\sum_{a \in A_t} |a| \overline{g}_a(q)^2}\right) = E\left(\sqrt{\sum_{a \in A_t} Z_a^2}\right), \tag{3.2}$$

where for cluster $a$,

$$\overline{g}_a(q) \equiv \frac{1}{|a|} \sum_{q \in a} g(q) \text{ and } Z_a \equiv \sqrt{|a|} \, \overline{g}_a(q).$$

The above holds since when $q \in a$ then $\frac{1}{\epsilon} \eta_t(q) = \overline{g}_a(q)$.

Note that $\{Z_a\}_{a \in A_t}$ are independent and identically distributed (iid) standard normal $N(0,1)$ rv's, since $\overline{g}_a(q) = N(0, 1/\sqrt{|a|})$. Thus letting $k = |A_t|$ (the number of clusters) and $\chi_k^2$ be a chi-square rv with $k$ degrees of freedom,

$$\frac{1}{\epsilon} E(\| \eta_t \|_2) = E\left(\sqrt{\sum_{a \in A_t} Z_a^2}\right) = E\left(\sqrt{\chi_k^2}\right) = \int_0^\infty dx \sqrt{x} \, \frac{1}{2^{k/2} \Gamma(k/2)} x^{k/2-1} e^{-x/2}$$

$$= \frac{1}{2^{k/2}} \frac{1}{\Gamma(k/2)} \int_0^\infty dx\, x^{(k-1)/2} e^{-x/2} = \frac{1}{2^{k/2}} \frac{1}{\Gamma(k/2)} \int_0^\infty 2\, dx\, (2x)^{(k-1)/2} e^{-x}$$

$$= \frac{2}{\sqrt{2}\Gamma(k/2)} \Gamma((k-1)/2+1) = \frac{\sqrt{2}}{\Gamma(k/2)} \Gamma(k/2+1/2)$$

using the density of the $\chi_k^2$ distribution in the third equality. Using the asymptotic series [33]:

$$\frac{\Gamma(j+1/2)}{\Gamma(j)} = \sqrt{j}\left(1 - \frac{1}{8j} + \frac{1}{128 j^2} + \frac{5}{1024 j^3} - \frac{21}{32768 j^4} + \cdots\right),$$

we obtain

$$\frac{1}{\epsilon} E(\|\eta_t\|_2) = \sqrt{2}\sqrt{k/2} + O(1/\sqrt{k}) = \sqrt{k} + O(1/\sqrt{k}).$$

(3.3)

Now

$$\frac{1}{\epsilon^2} V(\|\eta_t\|_2) = \frac{1}{\epsilon^2}\left[E(\|\eta_t\|_2^2) - E(\|\eta_t\|_2)^2\right] = E\left(\sum_{a \in A_t} Z_a^2\right) - \left(\sqrt{k} + O(1/\sqrt{k})\right)^2$$

$$= k - \left(\sqrt{k} + O(1/\sqrt{k})\right)^2 = O(1) \quad (k \to \infty),$$

(3.4)

completing the proof.

The following theorem can be used to denoise any function on a graph/network $G$ that varies 'slowly' relative to the network structure. The network plays the role of the Euclidean metric in Theorem 1, allowing denoising through local averaging. As above, $\|f\|^2 = \|f\|_2^2 = \int_G dq\, f(q)^2$.

We assume noisy measurements $f_1(q) = f(q) + \epsilon g(q)$ of $f(q)$, with $g(q)$ white noise on $G$ (an iid normal for each $q$). We use a filtration $\mathcal{F}_t$ on $G$, with the martingale $f_{1t} = E(f_1 \| \mathcal{F}_t)$ for denoising $f_1$. The following states Theorem 3 of the introduction more precisely.

**Theorem 3**: Consider all functions $f(q)$ on graphs $G$ that satisfy a uniformity condition

$$\|f_t - f\| - \|f_{t+1} - f\| \geq K(t) \qquad (3.5)$$

$(t = 0, 1, 2, \ldots, T)$ on the errors $\|f_t - f\|$ of their (finite) martingale approximations $f_t = E(f \| \mathcal{F}_t)$, with $K(t)$ a fixed positive function. Assume that the clusters $\{A_t\}_{0 \leq t \leq T}$ of the filter $\{\mathcal{F}_t\}_{0 \leq t \leq T}$ have cardinalities satisfying $|A_{t+1}| \geq C |A_t|$ for some $C > 1$. Let $f_{1t} = E(f_1 \| \mathcal{F}_t)$ be the martingale for recovering $f(q)$ from the perturbed function $f_1 = f(q) + \varepsilon(q)$. Then with a probability $p$ arbitrarily close to 1 if $1/\epsilon$ and $|G|$ are

sufficiently large, the approximation error $\| f_{1t} - f \|$ is non-monotonic, decreasing for small $t$ and increasing for large $t$, thus achieving a minimum for a positive value of the regularization parameter $t$.

**Remark.** As is shown below, the statement is uniform over all graphs $G$, all hierarchical clusterings $\{A_t\}_{0 \leq t \leq T}$ satisfying the growth condition on $|A_t|$, and all functions $f$ on $G$ satisfying (3.5), for fixed $K(t)$ and $C$. That is, the probability $p$ defined in the theorem converges to 1 uniformly as $\varepsilon \to 0$ and $|G| \mapsto \infty$.

**Proof:** Again letting $\eta(t) = \epsilon g(q)$ with the same definitions as in the Lemma, note that

$$\| \eta_{t+1} \| - \| \eta_t \| - \| f_{t+1} - f \| - \| f_t - f \| \leq \| f_{1(t+1)} - f \| - \| f_{1t} - f \|$$
$$\leq \| f_{t+1} - f \| - \| f_t - f \| + \| \eta_{t+1} \| + \| \eta_t \|, \tag{3.6}$$

since $f_{1t} - f = f_t + \eta_t - f$, with a similar statement for $f_t$. Thus by the Lemma (recall $k_t = |A_t|$),

$$\frac{1}{\epsilon}\left[ E(\| \eta_{t+1} \|) - E(\| \eta_t \|)\right] = \sqrt{|A_{t+1}|} - \sqrt{|A_t|} + O(1/\sqrt{|A_t|}) \geq \sqrt{C|A_t|} - \sqrt{|A_t|} + O(1/\sqrt{|A_t|})$$
$$= \left(\sqrt{C} - 1\right)\sqrt{|A_t|} + O(1/\sqrt{|A_t|}). \tag{3.7}$$

For random variables $X$ and $Y$, note

$$V(X+Y) \leq V(X) + V(Y) + 2\sqrt{V(X)V(Y)} \leq 2V(X) + 2V(Y).$$

Thus by (3.1),

$$\frac{1}{\epsilon^2}V(\| \eta_{t+1} \| - \| \eta_t \|) \leq \frac{2}{\epsilon^2}V(\| \eta_{t+1} \|) + \frac{2}{\epsilon^2}V(\| \eta_t \|) = O(1) \quad (|A_t| \mapsto \infty). \tag{3.8}$$

Since $f_t$ is an orthogonal projection of $f$, $\| f_t - f \| \leq \| f \|$, and

$$\| f_{1(t+1)} - f \| - \| f_{1t} - f \| \geq \| \eta_{t+1} \| - \| \eta_t \| - \| f_{t+1} - f \| - \| f_t - f \| \geq \| \eta_{t+1} \| - \| \eta_t \| - 2\| f \|,$$

and

$$P(\| f_{1(t+1)} - f \| - \| f_{1t} - f \| < 0) \leq P(\| \eta_{t+1} \| - \| \eta_t \| - 2\| f \| < 0) = P(B_t < 0),$$

where

$$B \equiv B_t \equiv \| \eta_{t+1} \| - \| \eta_t \| - 2\| f \|.$$

Note however that letting $\bar{B} \equiv E(B)$, (3.7) gives

$$\bar{B} = \epsilon\left(\sqrt{C}-1\right)\sqrt{|A_t|} + \epsilon O(1/\sqrt{|A_t|}) - 2\| f \| = \epsilon\left(\sqrt{C}-1\right)\sqrt{|A_t|} + O(1) \quad (|A_t| \mapsto \infty),$$

So that $\bar{B} > 0$ for large $t$. Note this statement is uniform over all graphs $G$ and admissible graph clusterings $\{A_s\}_{0 \leq s \leq T}$, (with $A_s$ the set of clusters at level $s$), so it is non-vacuous only for clusterings with $T \geq t$.

Thus

$$P(B < 0) = P(B - \bar{B} < -\bar{B}) \leq P(|B - \bar{B}| \geq \bar{B}) \leq \frac{V(B)}{\bar{B}^2}. \tag{3.9}$$

In addition, (3.8) gives
$$V(B) = V(\|\eta_{t+1}\| - \|\eta_t\|) = \epsilon^2 O(1) \quad (|A_t| \mapsto \infty).$$
Letting $L > 0$ be sufficiently large that the term $O(1) < L$ for all $t$, we have $V(B) \leq L\varepsilon^2$ for a universal constant $L$ (uniform over all graphs and cluster sizes $|A_t|$), so that

$$P(\|f_{1(t+1)} - f\| - \|f_{1t} - f\| < 0) \leq \frac{V(B)}{\overline{B}^2} = \frac{L}{\left[\left(\sqrt{C}-1\right)\sqrt{|A_t|} + \frac{O(1)}{\epsilon^2}\right]^2} \leq \frac{2L}{\left(\sqrt{C}-1\right)^2 |A_t|}, \quad (3.10)$$

for sufficiently large $|A_t|$, since $\epsilon$ is fixed.

Now note that for the sequence of clusterings (indexed by $t$), we have for a given $t_1$ sufficiently large,

$$P(B_t \leq 0 \text{ for some } t \geq t_1) \leq \sum_{t \geq t_1} P(B_t \leq 0) \leq \sum_{t \geq t_1} \frac{2L}{\left(\sqrt{C}-1\right)^2 |A_t|}.$$

Here we have required $t_1$ to be large enough for (3.10) to hold for all $t \geq t_1$. The sum on the right converges, since the $|A_t|$ grow geometrically. For large $t_1$ the right side and hence the left side above are arbitrarily small. So $P(B_t \leq 0 \text{ for some } t \geq t_1)$ is small for large $t_1$, uniformly over all graphs $G$ satisfying the conditions of the theorem. Thus for any $\delta > 0$, there is a $t_1 > 0$ such that $B_t$ is positive for all $t \geq t_1$ with probability greater than $1 - \delta$. Thus with probability greater than $1 - \delta$, we have $\|f_{1(t+1)} - f\| - \|f_{1t} - f\| > 0$ for all $t \geq t_1(\delta)$, assuming $G$ is large enough to accommodate a sequence of clusters with the sizes $A_t$. This completes the case of large $t$.

For small $t$, we have from (3.5)

$$\|f_{1(t+1)} - f\| - \|f_{1t} - f\| \leq \|f_{t+1} - f\| - \|f_t - f\| + \|\eta_{t+1}\| + \|\eta_t\| \leq \|\eta_{t+1}\| + \|\eta_t\| - K(t)$$

Note
$$E(\|\eta_{t+1}\| + \|\eta_t\| - K(t)) = E(\|\eta_{t+1}\|) + E(\|\eta_t\|) - K(t)$$
$$= \epsilon\left[\sqrt{|A_t|} + \sqrt{|A_{t+1}|} + O(|A_t|^{-1/2}) + O(|A_{t+1}|^{-1/2})\right] - K(t) < 0, \quad (3.11)$$

for $t \leq t_0$ (for any fixed $t_0$), and sufficiently small $\varepsilon < \varepsilon_0$ (with $\epsilon_0$ depending on $t_0$), since the $O(|A_t|^{-1/2})$ term is uniformly bounded in $t$ over $G$ and $|A_t|$. (Note there are only a finite number of integer values $t$ satisfying $0 \leq t \leq t_0$, and for all of them $K(t) > 0$). Furthermore by Lemma 2

$$V(\|\eta_{t+1}\| + \|\eta_t\| - K(t)) = V(\|\eta_{t+1}\|) + V(\|\eta_t\|) = \epsilon^2 O(1) \quad (t \to \infty).$$

where the $O(1)$ term can similarly be made uniform in the choice of $G$ and $|A_t|$.

Thus letting $D_t = \|\eta_{t+1}\| + \|\eta_t\| - K(t)$, we have (using the same Chebyshev inequality argument made in (3.9) (note $E(D) \equiv \overline{D} < 0$ by (3.11) for $\epsilon$ sufficiently small and $t \leq t_0$),

$$P(\| f_{1(t+1)} - f \| - \| f_{1t} - f \| \geq 0) \leq$$

$$P(D_t \geq 0) \leq \frac{V(D_t)}{\overline{D_t}^2} = \frac{\epsilon^2 O(1)}{\left\{ \epsilon \left[ \sqrt{|A_t|} + \sqrt{|A_{t+1}|} + O(|A_t|^{-1/2}) + O(|A_{t+1}|^{-1/2}) \right] - K(t) \right\}^2}, \quad (3.12)$$

which can be made arbitrarily small for $t \leq t_0$ for fixed $t_0$ if $\epsilon$ is sufficiently small. Putting together (3.10) and (3.12), it follows that for any $\delta > 0$, for sufficiently small $\varepsilon$ and large $|G| = |A_T|$, (recall the final clustering $A_T$ consists of singletons) with probability at least $1-\delta$, there exist $t_0 > 2$ and $t_1 > t_0$ such that for $t \leq t_0$, $\| f_{1t} - f \|$ is decreasing, while for $t \geq t_1$ it is increasing. (Note if $t_0 = 1$ monotonicity for $t \leq t_1$ is trivial). This statement is uniform over all finite graphs $G$ and clusterings $A_t$ satisfying $|A_{t+1}| \geq C |A_t|$ for a fixed $C$, and satisfying (3.5) for a fixed $K(t)$. This completes the proof.

We remark that above, generically, as $\epsilon$ decreases, the thresholds $t_0$ and $t_1$ both increase.

## 4. Kernel regression approximation: Euclidean spaces

Our second example of adapting regularization (smoothing) methods on $\mathbb{R}^d$ to regularizing feature vectors uses kernel regression. We will show first on $\mathbb{R}^d$ and then for the graph case, that optimal recovery typically means just the right amount of smoothing. Again in smoothing parameter $\alpha$, we are interested in when the optimal recovery is non-trivial, occurring at a finite positive regularization level $\alpha$. In this case the recovery error non-monotonic, worsening as $\alpha$ becomes very large and very small, with an intermediate $\alpha$ yielding optimal recovery.

For a function $f(x)$ on a domain $E \subset \mathbb{R}^d$ (with continuous boundary $\partial E$), suppose we are given a noisy version $f_1(q) = f(q) + \epsilon g(q)$, with (as earlier) $g(q)$ the Gaussian (white) noise distribution on $\mathbb{R}^d$ (see, e.g., [34]); note that $f_1(q)$ is a distribution since $g(q)$ is. We will recover an approximation of $f$ from $f_1(q)$ by smoothing, using standard Euclidean kernel regression to recover $f(q) \approx \sum_i \alpha_i K(q, z_i) + b$.

On $\mathbb{R}^d$ we assume a family of non-negative kernel functions $K_\alpha(q,r)$ ($q, r \in E$) parameterized by $\alpha > 0$, is an approximate identity, approaching the delta function $\delta(q-r)$ (i.e., the identity kernel) as $\alpha \to 0$. This means that (similarly to a Gaussian family $\frac{1}{(2\pi)^{d/2} \alpha^d} e^{-|q-r|^2/(2\alpha^2)}$) (1)

$$\int_R K_\alpha(q,r) d^n r \xrightarrow[\alpha \to 0]{} \begin{cases} 1 & \text{if } q \in H \\ 0 & \text{otherwise} \end{cases} \quad (4.1)$$

for any open $H \subset E$, (2) for $\alpha > 0$ $K_\alpha(q,r)$ is continuous in $\alpha$ and is bounded uniformly in $q,r$ by a continuous function of $\alpha$, and (3) that $K_\alpha(q,r)$ converges uniformly to 0 on $E \times E$ as $\alpha \to \infty$.

Through the regularization

$$f_1(q) \to K_\alpha f_1(q) \equiv \int_E dr\, K_\alpha(q,r) f_1(r) \equiv f_{1\alpha}(q),$$

we seek results analogous to those in Section 3, with kernel smoothing replacing local averaging as a denoising method.

We again seek conditions (based on the previous discussion) for a non-trivial regularization $f_{1\alpha}$ (with $\alpha > 0$) to improve the estimate of $f$ from $f_1$. That is, we seek a condition under which, for increasing values of the regularization $\alpha$, the error $\|f(q) - f_{1\alpha}(q)\|$ should first decrease and then increase, again yielding an intermediate $\alpha = \alpha_0$ for which error is minimized.

**Theorem 4:** Let $E$ be a closed bounded domain of $\mathbb{R}^d$, with continuous boundary, and $f(q)$ be a non-zero continuously differentiable function on $E$. Let $g(q)$ be standard Gaussian noise in $\mathbb{R}^d$, and define the perturbed function $f_1(q) = f(q) + \epsilon g(q)$. Let $K_\alpha(q,r)$ be an approximate identity, i.e., a family of bounded functions on $E \times E$, converging to the delta distribution $\delta(q-r)$ (as above) as $\alpha \to 0$. Then for small $\varepsilon$ the error $E(\|f_{1\alpha} - f\|_2)$ of the approximation $f_{1\alpha}(q) = \int_E K_\alpha(q,r) f_1(r) d^n r$ is non-monotonic in $\alpha \geq 0$ and this error is minimized at a positive regularization level $\alpha > 0$ (which depends on $\varepsilon$).

Again our interpretation is that error is minimized for positive $\alpha$, at a value that balances bias versus variance so that regularization improves estimation.

We define

$$G_\alpha(q) = \int_E K_\alpha(q,r) g(r) d^n r \text{ and } f_\alpha(q) = \int_E K_\alpha(q,r) f(r) d^n r \quad (4.2)$$

for the proof below.

We start with

**Lemma 1.** The expected norm $E(\|f_{1\alpha} - f\|)$ is continuous in $\alpha$.

**Proof.** Defining $R(\alpha) = E(\|f_{1\alpha} - f\|)$, we have

$$R(\alpha) - R(\alpha') = \| f_\alpha + \epsilon G_\alpha(q) - f\| - \|f_{\alpha'} + \epsilon G_{\alpha'}(q) - f\|$$
$$\leq \|(f_\alpha + \epsilon G_\alpha(q) - f) - (f_{\alpha'} + \epsilon G_{\alpha'}(q) - f)\| = \|f_\alpha - f_{\alpha'} + \varepsilon G_\alpha - \varepsilon G_{\alpha'}\|$$
$$\leq \|f_\alpha - f_{\alpha'}\| + \varepsilon \|G_\alpha - G_{\alpha'}\| \quad (4.3)$$

Taking expectations in (4.3),

$$|E(R(\alpha)) - E(R(\alpha'))| \leq \|f_\alpha - f_{\alpha'}\| + \varepsilon E(\|G_\alpha - G_{\alpha'}\|)$$

Now note that
$$G_\alpha(q) - G_{\alpha'}(q) = \int_E (K_\alpha(q,r) - K_{\alpha'}(q,r))g(r)dr$$
$$\sim N\left(0, \int (K_\alpha(q,r) - K_{\alpha'}(q,r))^2 dr\right) \sim b_q Z(q),$$

where $b_q^2 = \int (K_\alpha(q,r) - K_{\alpha'}(q,r))^2 dr$ and $Z(q)$ is a standard normal rv for each $q$. Thus $(G_\alpha(q) - G_{\alpha'}(q))^2 \sim b_q^2 \chi_1^2(q)$, with $\chi_1^2(q)$ a standard $\chi_1^2$ random variable for each $q$. Thus

$$\| G_\alpha - G_{\alpha'} \|^2 = \int_E b_q^2 \chi_1^2(q) dq,$$

and

$$E(\| G_\alpha - G_{\alpha'} \|^2) = \int_E b_q^2 E(\chi_1^2(q)) dq = \int_E b_q^2 dq$$
$$= \int_{E \times E} (K_\alpha(q,r) - K_{\alpha'}(q,r))^2 dr\, dq \underset{\alpha \to \alpha'}{\to} 0.$$

Hence it follows that $E(\| G_\alpha - G_{\alpha'} \|) \underset{\alpha \to \alpha'}{\to} 0$ since the expectation is over a (probability) space of measure 1. In addition, $f_\alpha(q) = \int_E f(r) K_\alpha(r,q) dr$ is continuous in $\alpha$, since

$$\| f_\alpha(q) - f_{\alpha'}(q) \|_q \leq \int_E \| K_\alpha(r,q) - K_{\alpha'}(r,q) \|_q f(r) dr \underset{\alpha \to \alpha'}{\to} 0,$$

by the dominated convergence theorem. Hence by (4.3) we conclude that $R(\alpha) - R(\alpha') \underset{\alpha' \to \alpha}{\to} 0$ in distribution, so that $E(R(\alpha))$ is continuous in $\alpha$, as desired.

**Lemma 2.** The expected norm $E(\| G_\alpha(q) \|) \underset{\alpha \to \infty}{\to} 0$.

**Proof.** By standard properties of white noise (as a multidimensional distribution $g(r)$, $r \in \mathbb{R}^n$),

$$\int_E K_\alpha(q,r) g(r) d^n r \equiv G_\alpha(q) \sim N\left(0, \sigma_\alpha^2(q) = \int_E K_\alpha^2(q,r) d^n r\right) \quad (4.4)$$

is a random variable-valued function of $q$. From (4.4), $G_\alpha(q)^2 \sim \sigma_\alpha^2(q) \chi_1^2(q)$, where $\chi_1^2(q)$ is a $\chi_1^2$ random variable for each $q$, and generally depends on $q$. Now note

$$E(\| G_\alpha \|^2) = \int_E E(G_\alpha^2(q)) dq = \int_E E(\sigma_\alpha^2(q) \chi_1^2(q)) dq$$

$$= \int_E \sigma_\alpha^2(q) E(\chi_1^2(q)) dq = \int_E \sigma_\alpha^2(q) dq = \| \sigma_\alpha^2(q) \|^2$$

$$= \int_E \int_E K_\alpha^2(q,r) d^n r\, d^n q \underset{\alpha \to \infty}{\to} 0$$

as desired, with the last limit depending on the above-mentioned uniform convergence of $K_\alpha(q,r)$ for large $\alpha$, and that $E$ is bounded set.

In the proof below $\|\cdot\|$ denotes an $L^2$ norm.

**Proof of Theorem 4:** We first consider the behavior of the expected error $E(R(\alpha))) \equiv E(\| f_{1\alpha} - f \|)$ for small $\alpha$. We write

$$f_{1\alpha}(q) = \int_E K_\alpha(q,r) f_1(r) d^n r = \int_E K_\alpha(q,r) f(r) d^n r + \epsilon \int_E K_\alpha(q,r) g(r) d^n r.$$
$$\equiv f_\alpha(q) + \epsilon G_\alpha(q), \qquad (4.5)$$

where $f_\alpha$ and $G_\alpha$ are as in (4.2). Note that

$$R(\alpha) \equiv \| f_{1\alpha} - f \| = \| f_\alpha + \epsilon G_\alpha(q) - f \| \geq \epsilon \| G_\alpha(q) \| - \| f_\alpha - f \|. \qquad (4.6)$$

For fixed $q$ note also that $G_\alpha(q)^2$ converges to $\infty$ in distribution as $\alpha \to 0$, since $\int_E K_\alpha^2(q,r) d^n r \underset{\alpha \to 0}{\to} \infty$. Hence also

$$\| G_\alpha(q) \|^2 = \int_{\mathbb{R}^d} G_\alpha^2(q) dq \underset{\alpha \to 0}{\to} \infty$$

in distribution, and so $\| G_\alpha(q) \|$ does the same, and therefore

$$E(\| G_\alpha(q) \|) \underset{\alpha \to 0}{\to} \infty. \qquad (4.7)$$

Since $\| f_\alpha - f \| \underset{\alpha \to 0}{\to} 0$, it follows from (4.6) and (4.7) that

$$E(R(\alpha)) \geq \varepsilon E(\| G_\alpha(q) \|) - \| f_\alpha - f \| \underset{\alpha \to 0}{\to} \infty,$$

so that $E(R(\alpha))$ increases unboundedly as $\alpha \to 0$

Now we consider the behavior of $E(R(\alpha))$ as $\alpha \to \infty$. In order to show that $E(R(\alpha))$ attains its minimum at a finite value of $\alpha$, we now show that it is greater than its minimum for large $\alpha > 0$. We now use the fact that

$$R(\alpha) \equiv \| f_{1\alpha} - f \| = \| f_\alpha + \epsilon G_\alpha(q) - f \| \geq \| f_\alpha - f \| - \epsilon \| G_\alpha(q) \|. \qquad (4.8)$$

Also by the uniform convergence to 0 for $K_\alpha(q,r)$ as $\alpha \to \infty$, it follows that $f_\alpha(q)$ converges (deterministically) to the zero function in $L^2(E)$. Thus $\| f_\alpha - f \| > \frac{1}{2} \| f \|$ for $\alpha$ large. By

$$E(R(\alpha)) \geq \| f_\alpha - f \| - \varepsilon E(\| G_\alpha(q) \|), \qquad (4.9)$$

if $\varepsilon > 0$ is sufficiently small then (for large $\alpha$)

$$\liminf_{\alpha \to \infty} E(R(\alpha)) \geq \lim_{\alpha \to \infty} \| f_\alpha - f \| - \varepsilon \limsup_{\alpha \to \infty} E(\| G_\alpha(q) \|)$$
$$> \frac{1}{2} \| f \| > 0.$$

On the other hand, for small $\varepsilon$ (i.e. as $\varepsilon \to 0$) the overall minimum value (over all $\alpha$)
$$\inf_{\alpha > 0} E(R(\alpha)) = \inf_{\alpha > 0} (\| f - f_\alpha \| + \varepsilon E(\| G_\alpha \|)) \qquad (4.10)$$
becomes arbitrarily small, given that $\| f_\alpha - f \| \underset{\alpha \to 0}{\to} 0$. In particular for small $\varepsilon$

$$\inf_{\alpha \geq 0} E(R(\alpha)) < \frac{1}{2} \| f \|, \tag{4.11}$$

the infimum being achieved (in $\alpha$) for $\alpha$ sufficiently close to 0 that the first term in (4.10) is small, while decreasing $\varepsilon > 0$ can make the second term arbitrarily small as well. In fact the infimum (4.11) can be made as close to 0 as desired by making $\varepsilon$ small enough, and in particular (4.11) will hold for small $\varepsilon$.

Combining the facts that $E(R(\alpha))$ is continuous in $\alpha$ (Lemma 2), increases to $\infty$ as $\alpha \to 0$, and (for $\varepsilon$ sufficiently small) has a minimum below $\frac{1}{2} \| f \|$, and increases above this value as $\alpha \to \infty$, it follows that (for sufficiently small $\varepsilon$), $\inf_{\alpha \geq 0} E(R(\alpha))$ is attained for a finite positive value of $\alpha$, i.e.,

$$0 < \arg\inf_{\alpha \geq 0} E(R(\alpha)) < \infty .$$

That is, a finite regularization level $\alpha_0 > 0$ achieves the minimal error $\| f_{1\alpha} - f \|$.

As before, this minimal error involves a compromise between positive bias $\| f_\alpha - f \|$ in exchange for diminished noise amplitude $\| G_\alpha(q) \|$.

## 5. Graph Kernel regression approximation theorem

We now extend Theorem 4 (which shows when kernel regularization improves approximation on $\mathbb{R}^d$), to the regularization of perturbed functions $f_1(q)$ on networks, again using kernel regression. As in Section 2, let $G$ be a network with a scalar-valued function $f$ defined on its nodes. We assume a perturbation of $f$ given by $f_1(q) = f(q) + \epsilon g(q)$, where $g(q)$ is Gaussian noise at each point in $G$, i.e., $g(q) \sim N(0,1)$ for each $q \in G$, with each $g(q)$ independent. We wish to regularize the noisy $f_1(q)$ by integrating it against a smoothing kernel $K_\alpha(q_1, q_2)$. We assume $K$ to be a non-negative-valued function with the property that

$$K_\alpha(q_1, q_2) \xrightarrow[\alpha \to 0]{} \delta(q_1, q_2) = \begin{cases} 1 & \text{if } q_1 = q_2 \\ 0 & \text{otherwise} \end{cases}$$

monotonically, in that as $\alpha \to 0$, $K_\alpha(q_1, q_2)$ increases if $q_1 \neq q_2$ and decreases if $q_1 = q_2$. We also assume that this convergence to the identity kernel occurs in the same way as in the previous section in $\mathbb{R}^n$, so $\sum_{q_1} K_\alpha(q, q_1) = 1$ and as $\alpha \to \infty$, $K_\alpha(q, q_1)$ converges to a fixed constant (which we assume without loss of generality is 0). We further assume $K_\alpha(q, q_1)$ is differentiable with respect to $\alpha$. We seek conditions parallel to those of Theorem 2 that guarantee that regularization can help, so there is a minimum of the error $\| f_{1\alpha} - f \|$ at a finite positive regularization parameter $\alpha$, intermediate between no regularization ($\alpha = 0$) and over-regularization ($\alpha \gg 1$).

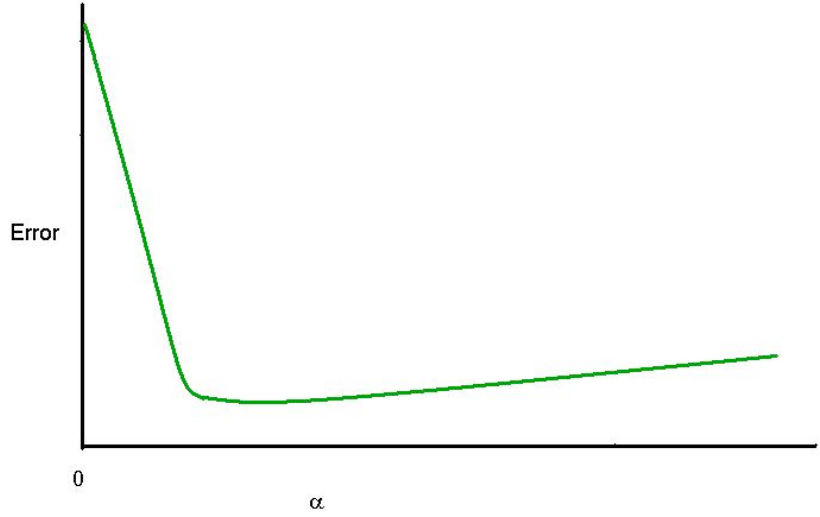

Figure 1: Schematic relationship between regularization parameter $\alpha$ and error of approximation for a perturbed noisy $f(q) = f_1(q) + \varepsilon g(q)$.

**Theorem 5**: Let $G$ be a graph or network, and $f(q)$ a function on $G$ that measured with error, yielding the noisy function $f_1(q) = f(q) + \epsilon g(q)$, with $g(q) \sim N(0,1)$ independnt standard Gaussian noise defined on $G$. Let the regression kernel $K_\alpha(q, q_1)$ converge to the identity operator (i.e., delta distribution) as above. Then for sufficiently small noise $\varepsilon > 0$, there is a positive finite regularization parameter $\alpha$ for which the expected error of the kernel regression approximation $f_{1\alpha}(q) = \sum_G f_1(q_1) K_\alpha(q_1, q) dq_1$ achieves a minimum. This error is non-monotonic, decreasing for small $\alpha$ and increasing for large $\alpha$.

To prove this we first show:

**Lemma 3.** Define the regularized functions

$$f_\alpha(q) = \sum_{q_1 \in G} K_\alpha(q, q_1) f(q_1); \quad g_\alpha(q) = \sum_{q_1 \in G} K_\alpha(q, q_1) g(q_1).$$

Then

$$E(\| f - f_{1\alpha} \|^2) = \sum_{q \in G} \left( f(q) - f_\alpha(q) \right)^2 + \epsilon^2 \sum_{q_1, q} K_\alpha(q, q_1)^2.$$

Proof: We have

$$\frac{d}{d\alpha} \| g_\alpha \|^2 = \frac{d}{d\alpha} \sum_q g_\alpha(q)^2 = \sum_q 2 g_\alpha(q) \frac{d}{d\alpha} g_\alpha(q)$$

and

$$E(\| f - f_{1\alpha} \|^2) = E(\| f - f_\alpha - \epsilon g_\alpha \|^2) = E\left[\sum_{q\in G}\left(f(q) - f_\alpha(q) - \epsilon g_\alpha(q)\right)^2\right]$$

$$= \sum_{q\in G}\left(f(q) - f_\alpha(q)\right)^2 + \epsilon^2 E\left(g_\alpha(q)^2\right) \qquad (4.12)$$

Also,

$$E(g_\alpha(q))^2 = E\left[\left(\sum_{q_1} K_\alpha(q,q_1) g(q_1)\right)^2\right] = E\left[\sum_{q_1} K_\alpha(q,q_1)^2 g(q_1)^2\right]$$

$$= \sum_{q_1} K_\alpha(q,q_1)^2 E\left(g(q_1)^2\right) = \sum_{q_1} K_\alpha(q,q_1)^2,$$

which increases as $\alpha \to 0$, and by (4.12)

$$E(\| f - f_{1\alpha} \|^2) = \sum_{q\in G}\left(f(q) - f_\alpha(q)\right)^2 + \epsilon^2 \sum_{q_1,q} K_\alpha(q,q_1)^2. \qquad (4.13)$$

as desired.

**Proof of Theorem 5:**

We will first show the error is decreasing for small $\alpha$. If $\alpha > 0$ is sufficiently small, we have by the Lemma

$$\frac{d}{d\alpha} E(\| f - f_{1\alpha} \|^2) = \frac{d}{d\alpha}\sum_{q\in G}\left(f(q) - f_\alpha(q)\right)^2 + \epsilon^2 \frac{d}{d\alpha}\sum_{q_1,q} K_\alpha^2(q,q_1).$$

$$= 2\sum_{q\in G}(f(q) - f_\alpha(q))\frac{d}{d\alpha} f_\alpha(q) + 2\epsilon^2 \sum_{q_1,q} K_\alpha(q,q_1)\frac{d}{d\alpha} K_\alpha(q,q_1)$$

$$= \sum_{q\in G} 2(f(q) - f_\alpha(q))\frac{d}{d\alpha}\sum_{q_1} K_\alpha(q,q_1) f(q_1) + 2\epsilon^2 \sum_{q_1,q} K_\alpha(q,q_1)\frac{d}{d\alpha} K_\alpha(q,q_1)$$

$$= \sum_{q,q_1\in G} 2(f(q) - f_\alpha(q)) f(q_1)\frac{d}{d\alpha} K_\alpha(q,q_1) + 2\epsilon^2 \sum_{q_1,q} K_\alpha(q,q_1)\frac{d}{d\alpha} K_\alpha(q,q_1)$$

$$= 2\sum_{q,q_1\in G}\left[(f(q) - f_\alpha(q)) f(q_1) + \epsilon^2 K_\alpha(q,q_1)\right]\frac{d}{d\alpha} K_\alpha(q,q_1)$$

$$= 2\sum_{q}\left[(f(q) - f_\alpha(q)) f(q) + \epsilon^2 K_\alpha(q,q)\right]\frac{d}{d\alpha} K_\alpha(q,q)$$

$$+2\sum_{q_1\neq q}\left[(f(q)-f_\alpha(q))f(q_1)+\epsilon^2 K_\alpha(q,q_1)\right]\frac{d}{d\alpha}K_\alpha(q,q_1)$$

$$\leq 2\sum_q \frac{1}{4}\epsilon^2 \frac{d}{d\alpha}K_\alpha(q,q)+2\sup_{q_1\neq q}\left|(f(q)-f_\alpha(q))f(q_1)+\epsilon^2 K_\alpha(q,q_1)\right|\sum_{q\neq q_1}\frac{d}{d\alpha}K_\alpha(q,q_1)$$

$$=\frac{1}{2}\epsilon^2\sum_q \frac{d}{d\alpha}K_\alpha(q,q)+2\sup_{q_1\neq q}\left|(f(q)-f_\alpha(q))f(q_1)+\epsilon^2 K_\alpha(q,q_1)\right|\sum_{q\neq q_1}\frac{d}{d\alpha}K_\alpha(q,q_1)$$

$$=\left[\left(\frac{1}{2}\epsilon^2-2\sup_{q_1\neq q}\left|(f(q)-f_\alpha(q))f(q_1)+\epsilon^2 K_\alpha(q,q_1)\right|\right)\sum_q \frac{d}{d\alpha}K_\alpha(q,q)\right]<0.$$

In the first inequality above (where the two sums are combined into one) we have used:

(a) $\frac{d}{d\alpha}K_\alpha(q,q)<0$ since $K_\alpha(q,q)$ increases as $\alpha\to 0$. Thus the coefficient $\frac{1}{4}\epsilon^2$ of the first sum on the right of the inequality needs to be less than $(f(q)-f_\alpha(q))f(q)+\epsilon^2 K_\alpha(q,q)$ for the inequality to hold.

(b) For $\alpha$ small, $K_\alpha(q,q)\geq \frac{1}{2}$, and

(c) $f(q)$ is uniformly bounded, and for $\alpha$ sufficiently small $f(q)-f_\alpha(q)$ is uniformly small in $q$, so $\left|(f(q)-f_\alpha(q))f(q_1)\right|\leq \frac{\epsilon^2}{4}$.

(d) $\frac{d}{d\alpha}K_\alpha(q,q_1)>0$ for $q\neq q_1$.

For the last equality we have used:

(a) $\sum_{q_1}K(q,q_1)=1$, so $\sum_{q_1\neq q}K_\alpha(q,q_1)=N-\sum_q K_\alpha(q,q)$ and

$\frac{d}{d\alpha}\sum_{q_1\neq q}K_\alpha(q,q_1)=-\frac{d}{d\alpha}\sum_q K_\alpha(q,q)$ with the sum on the left a double sum. Here $N=|G|$ is the size of the square matrix $K_\alpha(\cdot,\cdot)$

Finally, for the last inequality we have used:
(a) if $\alpha$ is sufficiently small, then the indicated supremum on the right is arbitrarily small.
(b) $\frac{d}{d\alpha}K_\alpha(q,q)<0$.

Since $\frac{d}{d\alpha} E(\| f - f_{1\alpha} \|^2) < 0$ at $\alpha = 0$, the minimum value of error occurs at $\alpha > 0$, as desired.

We now show that the above minimum occurs for a finite positive value of $\alpha$ (rather than at $\alpha = \infty$). First note that when $\varepsilon = 0$, the global minimum of

$$E(\| f - f_{1\alpha} \|^2) = \sum_{q \in G} \left( f(q) - f_\alpha(q) \right)^2 + \epsilon^2 \sum_{q_1, q} K_\alpha(q, q_1)^2 \qquad (4.14)$$

occurs at $\alpha = 0$. We will use a continuity argument to conclude that for sufficiently small $\varepsilon > 0$ the minimum of (4.14) is achieved at a finite $\alpha$. First, for some $\alpha_0$ (and for all $\varepsilon \geq 0$) the value of (4.14) for $\alpha > \alpha_0$ is strictly greater than some $\eta > 0$. However for sufficiently small $\varepsilon$, (4.14) is smaller than $\eta$ when $\alpha$ is close to 0. Thus for such (now fixed) $\varepsilon > 0$ (4.14) must achieve a minimum at some finite $\alpha \leq \alpha_0$. This minimum will not be at $\alpha = 0$, since (4.14) is decreasing at $\alpha = 0$. Hence for sufficiently small $\varepsilon > 0$ the minimum value for (4.14) is achieved at a finite positive $\alpha$, as desired.

## 6. Application: smoothed gene expression

In many current applications of machine learning, noise and so-called batch effects (originating from imprecise measurements or natural variations in measured populations) occur in data sets to be learned. These form a dominant problem preventing accurate learning and application of algorithms [1, 2]. In computational biology a standard technique for classification of tissue samples (e.g. to determine cancer subtypes) is based on a gene expression arrays, which measure expression levels $x_q$ of genes, usually numbering in the thousands to form a single feature vector $\mathbf{x}$ per tissue sample. Variations from different methodologies, along with abovementioned variations due to noise and batch effects in data, create difficulties at the data creation level in implementation of learning algorithms [3-5]. Because of such problems it is often considered adequate to have sensitivity and specificity of discriminations between two cancer tissue subtypes at 70-80% levels.

Mathematically the $i^{th}$ training tissue sample yields a feature vector $\mathbf{x}_i = (x_{i1}, \ldots, x_{in})$ (where $x_{iq}$ is the expression level of gene $q$ in sample $i$) and a class $y_i = 0$ or $1$, indicating which of the two cancer subtypes the training sample represents. This dataset of training vectors and classes forms a dataset $D = \{\mathbf{x}_i, y_i\}_{i=1}^m$. We view feature vector $\mathbf{x}_i = (x_{i1}, \ldots, x_{in})$ as a function $x_{iq} = f_i(q)$ on the feature index $q$. A machine learning algorithm uses information like that in $D$ to build a machine, a function $M$ which can take a novel sample's feature vector $\mathbf{x} = (x_1, \ldots, x_n)$ and classify its cancer subtype as $M(\mathbf{x}) = 0$ or $1$, predicting the subtype $y$. Given that measured feature vectors $\mathbf{x}_i$ are subject to noise, we seek to improve the vector by reducing noise, via smoothing of the feature function $f_i(q)$ defined on the genome, based on a network structure for genes. The gene network represents a similarity structure where genes are connected if they are likely to have similar expression levels, preferably based on prior information unrelated to the experiment producing $\mathbf{x}$. Here we use a so-called protein-protein interaction (PPI) network [35]. In such networks genes are connected if the proteins they code interact chemically. Thus regularization (denoising) of the feature vector $\mathbf{x}_i$ is accomplished by smoothing its feature function $f_i(q)$ over this network on genes $q$.

This is a pre-process entirely independent of the machine $M$; the regularization process improves the input data (training data $\mathbf{x}_i$ and the test data $\mathbf{x}$). The network gives prior information about a likely structure for $D$, providing input for regularizing feature vectors. This pre-processing step is implemented before any training or testing algorithms.

Such prior structural information has been used in various ways to eliminate error. In addition to the work of Rapaport [18] mentioned in the introduction, other methods that similarly smooth gene expression information in pre-processing have been developed. For prediction of breast cancer metastasis, [3-5] used opportunistic averaging of gene expression data over a protein-protein interaction gene network, selecting gene clusters based on their empirical predictive powers in a training set. In recent work on regularization/denoising of gene expression data, the re-interpretation of expression data as gene signatures averaged over pathway-based gene clusters has been studied more widely [36]

As mentioned earlier, the use of prior information (unrelated to current experimental data) for unsupervised regularization can improve feature vectors. The regularization can be done with prior graph or network structures on the genome (or analogous ones based on metric rather than network-based nearnesses). Incorporating such information may help to improve highly uncertain inference algorithms to have tolerable error levels.

We will consider two sets of gene expression microarray data from breast cancers, and use them as machine learning inputs to predict future tumor metastasis. Formally let $G = \{q_1, q_2, ..., q_n\}$ be a set of genes whose expression levels are measured. For this analysis we will assume each cancer has a 'true' underlying gene expression level $f(q)$, independent of noise, batch and methodology effects. We will incorporate the sum of the last three effects (which we collectively designate as 'noise') into a noise function $\eta(q)$. From the measured gene expression feature vector $\mathbf{x} = \{x_q\}_{q=1}^n$ (after all standard normalizations have been performed), with corresponding 'noisy' feature function $f_1(q) = f(q) + \eta(q)$, we seek a regularized function $f_{1\alpha}(q)$ and corresponding feature vector for implementation of ML algorithms to distinguish metastatic ($y=1$) from non-metastatic ($y=0$) subtypes.

For a dataset of subjects with breast cancer, consider the $i^{\text{th}}$ subject, and denote the true underlying expression value for genes $q_1, ..., q_n$ to be as $\mathbf{f}_i = (f_i(q_1), ..., f_i(q_n))$, so the values form a function on $G$. Thus in our model, $f_i(q)$ represents 'true' expression levels of genes (expression measurements that would be obtained without the above-mentioned collective noise). This ideal measurement is perturbed by a deviation $\eta_i(q)$ consisting of measurement noise. Henceforth index $i$ is omitted, and it is assumed the expression pattern $f(q)$ corresponds to a single subject (i.e., $i$ is fixed).

Thus (again suppressing $i$), gene $q \in G$ has expression $f_1(q) = f(q) + \eta(q)$ with noise $\eta(q) = \epsilon g(q)$ where $g(q)$ is standard Gaussian noise, assumed independent with distribution $N(0,1)$ for each $q$. Subject $i$ is thus represented as feature vector $\mathbf{f}_1 = (f_1(q_1), ..., f_1(q_n))$ of observed expression values. For subjects in the training group we also know whether their cancer will metastasize, and a label $y = y_i = 0$ or 1 indicates whether subject $i$ belongs to the metastasis group ($y = 1$ indicates metastasis). We seek to improve prediction via denoised expression values $\mathbf{f}_{1t}$, defined by averaging over clusters. This is done by taking the conditional expectation $f_{1t}(q) = E[f_1(q) | \mathcal{F}_t]$ of the measured gene expression function $\mathbf{f}_1 = f_1(q)$ with respect to $\sigma$-fields $\mathcal{F}_t$. These $\sigma$-fields are generated by hierarchically clustering the genome $G$ at different levels indexed by $t$. The clustering is based on a protein-protein interaction (PPI) network on $G$.
Thus the conditional expectations $f_{1t}(q)$ are simply averaged versions of the expression functions $f(q)$. More specifically, each $\sigma$-field $\mathcal{F}_t$ is generated by a disjoint partition (clustering) $\{G_k^{(t)}\}_k$ of the genome $G = \sqcup_k G_k^{(t)}$ for each value of the index $t$.

The conditional expectation $f_{1t}(q)$ of $f_1(q)$ on $G$ under $\mathcal{F}_t$ can re-defined as a function $\hat{f}_{1t}$ defined directly on the set of clusters $\{G_k^{(t)}\}_k$, so $\hat{f}_{1t}(G_k^{(t)}) = \sum_{q \in G_k^{(t)}} f_1(q) / |G_k^{(t)}|$, and $\hat{f}_{1t}$ is a piecewise averaging of the full expression function $f_1(q)$ defined on clusters $G_k^{(t)}$. As mentioned, biologically we assume genes in the same cluster should have similar expression, and thus yield more accurate values when stabilized by averaging over each cluster. For given $t$ this then gives an $n_t$-dimensional vector ($n_t$ is the number of gene clusters at level $t$) containing de-noised expression values, $\hat{\mathbf{f}}_{1t} = (\hat{f}_{1t}(G_1^{(k)}), ..., \hat{f}_{1t}(G_{n_t}^{(t)}))$. Such de-noised features can then be used in classification machine, in our case a support vector machine (SVM), in standard way.

**PPI network Clustering**
To test this approach a protein-protein interaction (PPI) network on the human genome $G$ is used here to identify (via graph clustering) protein subgroups that have similar or related functions. Thus neighboring proteins and those in a cluster are assumed to be in the same biochemical functional module, and thus to be co-expressed. We employ the *GraClus* software [37] to perform the graph clustering. A series of partitions $G^{(t)} = \sqcup_k G_k^{(t)}$ are then obtained for an increasing series of numbers of gene sets $n_t$.

**Combination of PPI network and Co-expression network**
A disadvantage of using the PPI network to cluster genes into groups with presumably similar expression patterns, is that such proteins, even if they function together, may in fact not have similar expression behavior. For example, a regulatory gene that inhibits its gene targets could exhibit a behavior which is inverse to that of its targets. Hence the conditional expectation above may allow gene pairs with opposite expressions to be averaged within the same cluster. To deal with this we can augment the above (un-weighted) PPI network $G$ to also incorporate information on co-expression correlations (i.e., correlations of expression patterns of gene pairs $q_i$ and $q_j$ across different subjects in the training data set). Specifically, an edge connecting the two genes can be weighted by $w_{ij} = \exp(-d_{ij}^2/\sigma^2)$, where $d_{ij}$ is a distance defined by hierarchical co-expression clustering of training data. The proteins most likely to be clustered together on this modified co-expression-adjusted PPI network $\{G, w\}$ do not only interact physically (based on their adjacency in the PPI network), but also are produced by genes which are manifestly co-expressed across patients, based on the above gene-gene correlations. The *GraClus* clustering algorithm can incorporate such weights in its clustering.

## Results
### Datasets
We have tested our algorithm on two breast cancer datasets from high-throughput gene expression studies by Wang, et al. [5] and van de Vijver, et al. [4] . In both datasets we have aimed to predict metastatic versus non-metastatic breast cancer patients based on their prior gene expression profiles **f** . The Wang dataset contains 286 breast cancer patients, of whom 93 eventually metastasized. The van de Vijver dataset contains 295 patients, of whom 79 metastasized.

The complete datasets contain expression measures of approximately 13,000 genes each. However in our expression feature vectors we use only genes contained in a PPI network that we compile from two databases, Reactome [38] and iRefIndex [39] . Thus 5,747 genes are kept in the Wang dataset (having a total of 70,353 documented PPI interactions), and 5,310 genes (with 67,342 interactions) for the van't Vijver dataset.

### Experimental Protocols
In order to test the machine learning algorithms (SVM) trained on these data, we use a 5-fold cross-validation, in which the original datasets are divided into 5 groups (folds) of equal size, and training of the machine classifier is done using 4 of these groups, with testing of the classifier performance on the 5$^{th}$ group. Thus a machine model is trained on the 4 folds (used as training data, in which metastasis outcomes $y_i$ are known), and then used to score each reserved patient in the 5$^{th}$ fold to predict metastasis or no metastasis. The training and test groups are rotated through the 5 folds until all 5 groups have been used 4 times each as training data and once as test data. Three measures are used to compare the performance of the above-mentioned prediction algorithms, including area under the ROC curve (AUROC) and the area under the precision-recall curve (AUPRC). The ROC curve is a plot of sensitivity versus specificity, while the precision-recall curve is a plot of precision versus recall (sensitivity). Both curves assess the classification performance by balancing the type I and type II errors. Note that in this 5-fold protocol, the clusters are determined only by the 4 training folds, and only these clusters are used in training and in testing. In the case of just the PPI network, we simply build fixed (data-independent) clusters without any cross-validation protocol, since the clustering procedure does not use any information from the label $y_i$ .

**Use of cluster data improves classification performance**

We test the three aforementioned clustering methods on each breast cancer dataset. We choose the numbers of clusters $n_t$ to be 64, 128, 256, 512 1024 and 2048 (so that the above clustering parameter $t$ takes on 6 values). The machine learning predictive accuracies on both cancer metastasis datasets are improved compared with the same methods using individual gene features (i.e., for which $n_t > 5000$). For example, the area under the precision recall curve is improved by co-expression-adjusted PPI clustering from 36.2% to 52.4% and from 34.6% to 43.0% for the Wang and van de Vijver datasets, respectively. Details are listed in **Table 1a** and **Table 1b**.

| PPI + Expr | $k$ = 64 | 128 | 256 | 512 | 1024 | 2048 | All Genes |
|---|---|---|---|---|---|---|---|
| AUROC | 0.668(0.013) | 0.691(0.015) | 0.714(0.017) | 0.705(0.018) | 0.724(0.021) | **0.732(0.018)** | 0.534(0.044) |
| AUPRC | 0.466(0.021) | 0.489(0.025) | 0.511(0.029) | 0.491(0.029) | 0.512(0.034) | **0.524(0.029)** | 0.362(0.032) |

**Table 1a**: The performance of co-expression adjusted PPI network clustering on Wang's dataset. Quantities in parentheses represent standard deviations over 100 trials (random selections of 5 folds).

| PPI + Expr | $k$ = 64 | 128 | 256 | 512 | 1024 | 2048 | All Genes |
|---|---|---|---|---|---|---|---|
| AUROC | 0.690(0.015) | 0.708(0.013) | 0.721(0.015) | 0.715(0.018) | 0.711(0.018) | **0.729(0.017)** | 0.660(0.027) |
| AUPRC | 0.385(0.019) | 0.401(0.019) | 0.421(0.022) | 0.402(0.022) | 0.404(0.025) | **0.430(0.027)** | 0.346(0.028) |

**Table 1b**: The performance of co-expression adjusted PPI network clustering on van de Vijver's dataset.

We observe that that as $n_t$ increases classification performance improves until an optimal number (between 1024 and 2048) is reached, consistent with the statement of Theorem 2. We thus average gene expressions at increasingly refined (small cluster) levels until approaching the unaggregated individual gene expression levels, which then lowers prediction performance. As indicated in Theorem 2, averaging over too small a gene cluster does not sufficiently quench noise $\eta(q)$. We observe also that co-expression adjustment improves performance (see **Table 2a** and **Table 2b**), suggesting that such aggregated feature vectors do the best job of quenching noise in gene expression arrays.

| PPI | $k$ = 64 | 128 | 256 | 512 | 1024 | 2048 | All Genes |
|---|---|---|---|---|---|---|---|
| AUROC | 0.658(0.014) | 0.680(0.015) | 0.692(0.019) | 0.684(0.019) | 0.708(0.019) | **0.730(0.017)** | 0.534(0.044) |
| AUPRC | 0.450(0.019) | 0.462(0.021) | 0.475(0.026) | 0.487(0.031) | 0.500(0.032) | **0.522(0.029)** | 0.362(0.032) |

**Table 2a**: Performance of plain PPI network clustering on Wang's dataset.

| PPI | $k$ = 64 | 128 | 256 | 512 | 1024 | 2048 | All Genes |
|---|---|---|---|---|---|---|---|
| AUROC | 0.687(0.014) | 0.705(0.013) | 0.689(0.016) | 0.686(0.021) | **0.712(0.019)** | 0.503(0.035) | 0.660(0.027) |
| AUPRC | 0.371(0.015) | 0.399(0.019) | 0.398(0.022) | 0.375(0.023) | **0.403(0.026)** | 0.270(0.026) | 0.346(0.028) |

**Table 2b**: Performance of plain PPI network clustering on van de Vijver's dataset.

We have also tested the performance of random clustering (**Table 3a** and **Table 3b**), in which genes are clustered randomly into $n_t$ clusters, independently of graph $G$ or other biological information. For random clustering, performance improves as well, which can be explained by the fact that the noise $\eta(q)$ can still be quenched by the smoothing over relatively large clusters, including random ones. As expected, true underlying expression cannot be as accurately approximated as in biology-based clustering, given randomly grouped genes lack the same underlying expression behavior. As shown in **Tables 3a and 3b)**, the optimal numbers of clusters for random clustering are smaller than for meaningful (non-random) clustering.

| **Random** | $k = 64$ | 128 | 256 | 512 | 1024 | 2048 | All Genes |
|---|---|---|---|---|---|---|---|
| AUROC | 0.556(0.031) | 0.545(0.030) | 0.600(0.032) | **0.617(0.029)** | 0.500(0.035) | 0.502(0.034) | 0.534(0.044) |
| AUPRC | 0.409(0.013) | 0.433(0.019) | 0.487(0.025) | **0.514(0.032)** | 0.341(0.026) | 0.338(0.026) | 0.362(0.032) |

**Table 3a**: The performance of random clustering on Wang's dataset.

| **Random** | $k = 64$ | 128 | 256 | 512 | 1024 | 2048 | All Genes |
|---|---|---|---|---|---|---|---|
| AUROC | 0.579(0.036) | 0.617(0.033) | **0.640(0.030)** | 0.627(0.033) | 0.509(0.039) | 0.504(0.038) | 0.660(0.027) |
| AUPRC | 0.340(0.015) | 0.381(0.017) | 0.378(0.019) | **0.426(0.028)** | 0.271(0.027) | 0.272(0.023) | 0.346(0.028) |

**Table 3b**: The performance of random clustering on van de Vijver's dataset.

Though this expression feature preprocessing method produces more accurate classification, a strict performance assessment for this as an application in computational biology would also compare performance of our preprocessing algorithm with un-preprocessed performance, but including feature selection methods on both. We have noted that the performance of regularization followed by dimensional reduction is comparable to just dimensional reduction, using SVMRFE [40] as a feature selector. However, to distinguish the two methodologies our feature vector regularization is fully unsupervised, i.e., does not use (known or unknown) data classes $y_i$. Hence it plays a role that is orthogonal to standard supervised feature selection methods, which depend on knowing the classes $y_i$ of all feature vectors $\mathbf{x}_i$. This unsupervised regularization method can in particular be used either before or after supervised feature selection, or without it. Being only dependent on the data $\mathbf{x}_i$ and not the classes, the regularization method is independent of the machine $M$ that is subsequently trained and used on the data. Since the method applies independently of dimensional reductions, we expect its benefits to be supplemental to those of standard dimensional reduction methods, though this is not the case here. The processes of unsupervised regularization and feature selection are independent and deal with different stages of classifier-building.

**Bibliography:**

[1] G. Sanguinetti, M. Milo, M. Rattray, and N. D. Lawrence, "Accounting for probe-level noise in principal component analysis of microarray data," *Bioinformatics,* vol. 21, pp. 3748-3754, 2005.

[2] V. Quaranta and S. P. Garbett, "Not all noise is waste," *nature methods,* vol. 7, pp. 269-272, 2010.

[3] E. Lee, H. Y. Chuang, J. W. Kim, T. Ideker, and D. Lee, "Inferring pathway activity toward precise disease classification," *PLoS Computational Biology,* vol. 4, p. e1000217, 2008.

[4] M. J. Van De Vijver, Y. D. He, L. J. van't Veer, H. Dai, A. A. M. Hart, D. W. Voskuil, G. J. Schreiber, J. L. Peterse, C. Roberts, and M. J. Marton, "A gene-expression signature as a predictor of survival in breast cancer," *New England Journal of Medicine,* vol. 347, pp. 1999-2009, 2002.

[5] Y. Wang, J. G. M. Klijn, Y. Zhang, A. M. Sieuwerts, M. P. Look, F. Yang, D. Talantov, M. Timmermans, M. E. Meijer-van Gelder, and J. Yu, "Gene-expression profiles to predict distant metastasis of lymph-node-negative primary breast cancer," *The Lancet,* vol. 365, pp. 671-679, 2005.

[6] T. Hastie, R. Tibshirani, J. Friedman, and J. Franklin, "The elements of statistical learning: data mining, inference and prediction," *The Mathematical Intelligencer,* vol. 27, pp. 83-85, 2005.

[7] V. N. Vapnik, "Statistical learning theory," 1998.

[8] P. Niyogi, F. Girosi, and T. Poggio, "Incorporating prior information in machine learning by creating virtual examples," *Proceedings of the IEEE,* vol. 86, pp. 2196-2209, 1998.

[9] S. G. Chang, B. Yu, and M. Vetterli, "Adaptive wavelet thresholding for image denoising and compression," *Image Processing, IEEE Transactions on,* vol. 9, pp. 1532-1546, 2000.

[10] J.-L. Starck, E. J. Cand√®s, and D. L. Donoho, "The curvelet transform for image denoising," *Image Processing, IEEE Transactions on,* vol. 11, pp. 670-684, 2002.

[11] T. F. Chan, S. Esedoglu, and M. Nikolova, "Algorithms for finding global minimizers of image segmentation and denoising models," *SIAM Journal on Applied Mathematics,* vol. 66, pp. 1632-1648, 2006.

[12] A. N. Tikhonov, "Stable methods for the summation of Fourier series," *Soviet Math. Dokl.,* vol. 5, p. 4, 1964.

[13] N. M. Krukovskii, "On the Tikhonov-stable summation of Fourier series with perturbed coefficients by some regular methods," *Moscow Univ. Math. Bull.,* vol. 28, p. 7, 1973.

[14] S. Kim, M. Kon, and C. DeLisi, "Pathway-based classification of cancer subtypes," *Biology Direct,* vol. 7, p. 21, 2012.

[15] M. Liu, A. Liberzon, S. W. Kong, W. R. Lai, P. J. Park, I. S. Kohane, and S. Kasif, "Network-based analysis of affected biological processes in type 2 diabetes models," *PLoS genetics,* vol. 3, p. e96, 2007.

[16] D. K. Hammond, P. Vandergheynst, and R. Gribonval, "Wavelets on graphs via spectral graph theory," *Applied and Computational Harmonic Analysis,* vol. 30, pp. 129-150, 2011.

[17] A. Smola and R. Kondor, "Kernels and regularization on graphs," *Learning theory and kernel machines,* pp. 144-158, 2003.

[18] F. Rapaport, A. Zinovyev, M. Dutreix, E. Barillot, and J.-P. Vert, "Classification of microarray data using gene networks," *BMC Bioinformatics,* vol. 8, p. 35, 2007.

[19] A. D. Szlam, M. Maggioni, and R. R. Coifman, "Regularization on graphs with function-adapted diffusion processes," *The Journal of Machine Learning Research,* vol. 9, pp. 1711-1739, 2008.

[20] M. Belkin and P. Niyogi, "Laplacian eigenmaps for dimensionality reduction and data representation," *Neural computation,* vol. 15, pp. 1373-1396, 2003.

[21] S. Bougleux, A. Elmoataz, and M. Melkemi, "Discrete regularization on weighted graphs for image and mesh filtering," *Scale Space and Variational*

*Methods in Computer Vision,* pp. 128-139, 2007.

[22] S. Sardy, A. G. Bruce, and P. Tseng, "Block coordinate relaxation methods for nonparametric wavelet denoising," *Journal of computational and graphical statistics,* vol. 9, pp. 361-379, 2000.

[23] R. R. Coifman and D. L. Donoho, "Translation-invariant de-noising," *LECTURE NOTES IN STATISTICS-NEW YORK-SPRINGER VERLAG-,* pp. 125-125, 1995.

[24] A. Buades, B. Coll, and J. M. Morel, "A non-local algorithm for image denoising," in *Computer Vision and Pattern Recognition, 2005. CVPR 2005. IEEE Computer Society Conference on*, 2005, pp. 60-65.

[25] A. J. Smola and B. Scholkopf, "A tutorial on support vector regression," *Statistics and computing,* vol. 14, pp. 199-222, 2004.

[26] S. Geman, E. Bienenstock, and R. Doursat, "Neural networks and the bias/variance dilemma," *Neural computation,* vol. 4, pp. 1-58, 1992.

[27] W. Hardle, G. Kerkyacharian, D. Picard, and A. Tsybakov, *Wavelets, approximation, and statistical applications*: Springer New York, 1998.

[28] F. Cucker and S. Smale, "Best choices for regularization parameters in learning theory: On the bias-variance problem," *Foundations of Computational Mathematics,* vol. 2, pp. 413-428, 2002.

[29] T. Poggio and S. Smale, "The mathematics of learning: Dealing with data," *Notices of the AMS,* vol. 50, pp. 537-544, 2003.

[30] W. Feller, "Introduction to Probability Theory and Its Applications, Vol. II POD," 1974.

[31] Y. Fan, M. Kon, S. Kim, L. Raphael, and C. DeLisi, "Regularization techniques for machine learning on graphs and networks with biological applications," *Communications in Mathematical Analysis,* vol. 8, pp. 136-145, 2010.

[32] Y. Fan, S. Kim, M. Kon, L. Raphael, and C. DeLisi, "Smoothing gene expression using biological networks," *Machine Learning and Applications,* vol. 9, pp. 540-545, 2010.

[33] R. Graham, D. Knuth, and O. Patashnik, "Answer to problem 9.60 in concrete mathematics: A foundation for computer science," ed: Reading, MA: Addison-Wesley, 1994.

[34] H. Holden, B. √òksendal, J. Ub √ ∏e, and T. Zhang, *Stochastic partial differential equations: a modeling, white noise functional approach*: Birkhauser Boston Inc., 1996.

[35] J.-F. o. Rual, K. Venkatesan, T. Hao, T. Hirozane-Kishikawa, A. l. Dricot, N. Li, G. F. Berriz, F. D. Gibbons, M. Dreze, and N. Ayivi-Guedehoussou, "Towards a proteome-scale map of the human protein‚Äìprotein interaction network," *Nature,* vol. 437, pp. 1173-1178, 2005.

[36] C. J. Vaske, S. C. Benz, J. Z. Sanborn, D. Earl, C. Szeto, J. Zhu, D. Haussler, and J. M. Stuart, "Inference of patient-specific pathway activities from multi-dimensional cancer genomics data using PARADIGM," *Bioinformatics,* vol. 26, pp. i237-i245, 2010.

[37] I. S. Dhillon, Y. Guan, and B. Kulis, "Weighted graph cuts without eigenvectors a multilevel approach," *Pattern Analysis and Machine Intelligence,*

*IEEE Transactions on,* vol. 29, pp. 1944-1957, 2007.

[38] G. Joshi-Tope, M. Gillespie, I. Vastrik, P. D'Eustachio, E. Schmidt, B. de Bono, B. Jassal, G. Gopinath, G. Wu, and L. Matthews, "Reactome: a knowledgebase of biological pathways," *Nucleic Acids Research,* vol. 33, pp. D428-D432, 2005.

[39] S. Razick, G. Magklaras, and I. M. Donaldson, "iRefIndex: a consolidated protein interaction database with provenance," *BMC Bioinformatics,* vol. 9, p. 405, 2008.

[40] I. Guyon and A. Elisseeff, "An introduction to variable and feature selection," *The Journal of Machine Learning Research,* vol. 3, pp. 1157-1182, 2003.